\documentclass{article}

\usepackage[dblblindworkshop, preprint]{neurips_2026}
\workshoptitle{AI4Mat}
\usepackage{amsthm}
\newtheorem{proposition}{Proposition}
\usepackage[utf8]{inputenc}
\usepackage[T1]{fontenc}
\usepackage{amsmath,amssymb,mathtools}
\usepackage{chngcntr}
\usepackage{caption}  
\usepackage{booktabs}
\usepackage{graphicx}
\usepackage{xcolor}
\usepackage{pgfplots}
\pgfplotsset{compat=1.18}
\usepackage{microtype}
\usepackage{hyperref}
\usepackage{url}
\usepackage{float}
\usepackage{color-edits}
\addauthor{ry}{purple}
\addauthor{pb}{blue}
\addauthor{pf}{red}
\addauthor{ma}{orange}
\addauthor{sg}{green}

\newcommand{\R}{\mathbb{R}}

\title{Investigating Composite Modeling in Multi-Objective Bayesian Optimization for Scientific Design}
\author{%
  Ricky Ye$^{1}$\thanks{Equal contribution.}
  \quad Sahas Goli$^{1}$\footnotemark[1]
  \quad Muhammad Ali$^{1}$\footnotemark[1]
  \quad Poompol Buathong$^{2}$
  \quad Peter I.~Frazier$^{3}$
  \\[0.7em]
  \normalfont\small
  \begin{tabular}{c}
    $^{1}$Department of Computer Science, Cornell University\\
    $^{2}$Center for Applied Mathematics, Cornell University\\
    $^{3}$School of Operations Research and Information Engineering, Cornell University\\[0.4em]
    \texttt{\{%
\href{mailto:rjy42@cornell.edu}{rjy42},%
\href{mailto:sg2636@cornell.edu}{sg2636},%
\href{mailto:ma2379@cornell.edu}{ma2379},%
\href{pb482@cornell.edu}{pb482},%
\href{mailto:pf98@cornell.edu}{pf98}%
\}@cornell.edu}
  \end{tabular}%
}

\begin{document}
\maketitle
\begin{abstract}
Scientific design often requires balancing competing objectives that are expensive to evaluate through experiments or simulations. Multi-objective Bayesian optimization (MOBO) addresses this challenge by using probabilistic models to guide the selection of promising designs. In many applications, tunable inputs map to observable intermediate responses, which are then combined through known functions to compute the final objectives. Prior work on Bayesian optimization of composite functions shows that learning from these intermediate responses, which we call Composite modeling, can substantially improve sampling efficiency in single-objective optimization, compared to methods that condition only on black-box observations of the objective function. The benefits of Composite modeling in MOBO, however, remain understudied. In this work, we study the benefits of Composite modeling in MOBO and investigate the factors driving variation in its gains. 
We first establish single-objective conditions under which intermediate observations provide no additional predictive information beyond objective values, where composite and black box methods use the same warped Gaussian process prior. Controlled numerical experiments with versions of four MOBO methods across eight synthetic and application-inspired scientific-design benchmarks examine how these conditions relate to optimization gains in the multi-objective setting, where the black box approach models the objective functions with a standard multi-output Gaussian process.
Both investigations support the claim that composite modeling tends to help more when the function mapping intermediate responses to final objectives is non-bijective and non-affine.
\end{abstract}
\section{Introduction}
\label{sec:introduction}

Scientific design often involves balancing competing goals. A material, for example, may need to perform well while also being inexpensive. Evaluating a design can require a costly experiment. We formulate this problem as
$\min_{\mathbf{x}\in\mathcal{X}} \mathbf{f}(\mathbf{x})=(f_1(\mathbf{x}),\ldots,f_N(\mathbf{x}))$, where
$\mathcal{X}\subseteq\mathbb{R}^d$ is the design space.
Multi-objective Bayesian optimization (MOBO) seeks favorable tradeoffs
with few evaluations~\cite{gopakumar2018multiobjective,knowles2006parego},
using Gaussian process (GP) models~\cite{rasmussen2006gaussian} and an
acquisition function to select designs.

In many scientific applications, tunable inputs map to observable intermediate responses, which are then combined through known functions to compute the final objectives. These objectives have the composite form
$f_i(\mathbf{x})=g_i\bigl(\mathbf{h}_i(\mathbf{x})\bigr)$ for $i=1,\ldots,N$
where \(\mathbf{h}_i:\mathcal{X}\rightarrow\mathbb{R}^{k_i}\) is an expensive black-box function returning \(k_i\) intermediate responses and \(g_i:\mathbb{R}^{k_i}\rightarrow\mathbb{R}\) is known and inexpensive to evaluate. For single objectives, we write $f,g$ and $\mathbf{h}$ in place of $f_1, g_1$ and $\mathbf{h}_1$.
For example, in color formulation, dye concentrations determine
measured color coordinates, which known functions convert into
color differences from a target~\cite{ginsburg2023color}.
Matching under different lighting can create competing
objectives~\cite{chaoucha2022color}.

Bayesian optimization (BO) of composite functions~\cite{astudillo2019bayesian}
models intermediate responses and propagates posterior samples through
the known mappings. This can substantially improve single-objective
sampling efficiency, with extensions to high-dimensional outputs
and function networks~\cite{maddox2021outputs,maddox2021physics,
astudillo2021bayesian,buathong2024bayesian}. Composite modeling has
also been applied to MOBO through smooth Tchebycheff
scalarization (STCH)~\cite{lin2024smooth,pires2026composite}, but the factors
governing its benefits remain understudied.

We investigate the benefits of composite modeling through three inference approaches.
\emph{Composite} places a GP prior on the intermediate responses and
conditions on their observations. \emph{BlackBox-GP} places a GP prior
directly on the objectives and conditions on objective observations.
\emph{BlackBox-Induced} uses the objective prior induced by Composite's
prior through the known mappings, but conditions only on objective
observations. 
BlackBox-GP corresponds directly to how BO is usually implemented in practice while BlackBox-Induced supports a precise comparison of the benefits provided by the additional information provided by intermediate responses while holding the prior distribution fixed.


We find evidence that composite modeling tends to perform as well or better than black box approaches and provides the largest benefits when the functions $g_i$ are non-bijective and non-affine.
First, we analyze the single-objective case and show that Composite and BlackBox-Induced create equivalent posterior distributions when $g$ is bijective or when $g$ is affine and the prior on $\mathbf{h}$ has separable covariance. This suggests composite approaches are equal to black box ones in these settings. 
Second, experiments comparing Composite with BlackBox-GP on both single and multi-objective problems show that Composite provides a positive mean final-hypervolume gain in 15 of 16 comparisons and that the problems where Composite provides the biggest benefits are non-affine and non-bijective $g_i$.

\section{Problem Setting}

Consider objectives $f_i(\mathbf{x})=g_i(\mathbf{h}_i(\mathbf{x}))$, $i=1,\ldots,N$,
where $\mathbf{h}_i:\mathcal{X}\rightarrow\mathbb{R}^{k_i}$ is unknown
and $g_i:\mathbb{R}^{k_i}\rightarrow\mathbb{R}$ is known.
The intermediate responses may be shared across objectives.
We consider both single-objective ($N=1$) and multi-objective
($N>1$) settings. For the single-objective analysis, we omit
the subscript and write $f(\mathbf{x})=g(\mathbf{h}(\mathbf{x}))$.


We consider three ways of performing inference in this setting, one that conditions on observations of $\mathbf{h}(\mathbf{x})$ and two that condition only on observations of $f(\mathbf{x})$.
The first method, called {\it Composite}, aligns with composite BO~\cite{astudillo2019bayesian}: we put a multi-output GP on $\mathbf{h}(\mathbf{x})$ and condition on observations of $\mathbf{h}(\mathbf{x})$.
The second method, called {\it BlackBox-GP}, aligns with classical BO~\cite{jones1998efficient,frazier2018bayesian}: we put a GP on $f(\mathbf{x})$ and condition on observations of $f(\mathbf{x})$. This treats $f(\mathbf{x})$ as a black box, supposing that we cannot observe the inner function $\mathbf{h}(\mathbf{x})$.
The third method, called {\it BlackBox-Induced}, supports theoretical
analysis: we start with the same GP prior on $\mathbf{h}(\mathbf{x})$ as Composite,
induce a corresponding prior on $f(\mathbf{x})$ through the known mapping
$g$, and condition on observations of $f(\mathbf{x})$.  Thus, this method uses the same prior as Composite but conditions on the same black box information as BlackBox-GP.
This is a form of warped GP prior \citep{snelson2003warped} with a fixed warping function. Appendix~\ref{app:induced} describes how to perform posterior inference under this induced prior.

Below, using theory and experiments, we analyze whether observing $\mathbf{h}(\mathbf{x})$ (in the Composite method) provides a benefit over observing only $f(\mathbf{x})$ (the two BlackBox methods) and how this depends on the choice of $g$. We focus specifically on whether $g$ is a bijection or is affine, finding support for the claim that conditioning on $\mathbf{h}(\mathbf{x})$ does {\it not} provide a benefit when $g$ is a bijection or is affine, and often {\it does} provide a benefit when $g$ is neither bijective nor affine.

\section{Theoretical Analysis}
We first compare Composite and BlackBox-Induced in a single-objective case theoretically.

\begin{proposition}[Posterior equivalence]
\label{prob:equivalence}
At fixed evaluation locations $\mathbf{x}_1,\ldots,\mathbf{x}_n$, let $
H=(\mathbf{h}(\mathbf{x}_1),\ldots,\mathbf{h}(\mathbf{x}_n)),$ and $
F=(f(\mathbf{x}_1),\ldots,f(\mathbf{x}_n)).$ Suppose $\mathbf{h}$ is drawn from a GP prior with fixed known mean function and kernel.
Suppose also that either (1) $g$ is bijective on the possible intermediate responses, so each objective value uniquely determines its intermediate response; or (2) $g(\mathbf{z})=\mathbf{a}^\top \mathbf{z}+b$ for some fixed $\mathbf{a}$ and $b$, and the GP prior on $\mathbf{h}$ has separable covariance.
Then, for every fixed $\mathbf{x}\in\mathcal{X}$
and $t\in\mathbb{R}$,
$
\mathbb{P}(f(\mathbf{x})\leq t\mid H)
=
\mathbb{P}(f(\mathbf{x})\leq t\mid F)
\text{ with probability one}.
$
\end{proposition}

Technical conditions and the proof are provided in Appendix~\ref{app:proof-posterior-equivalence}. Under the Proposition~\ref{prob:equivalence}, which applies to \textit{Identity} and \textit{Cubic} experiments below, Composite and BlackBox-Induced yield the same posterior over $f$ when they share a GP prior on $\mathbf{h}$. 
Consequently, acquisition rules determined solely by the posterior produce the same decisions and final posterior distribution.
In these problems, 
differences observed between Composite and BlackBox-GP are due to the prior distribution, Monte Carlo noise, or small implementation differences, since BlackBox-GP and BlackBox-Induced condition on the same information.

\section{Experimental Analysis}

\begin{table}[t]
    \centering
    \small
    \setlength{\tabcolsep}{3pt}
        \caption{
        Controlled synthetic experiments. Final regret reductions (single-objective) and final hypervolume gains (multi-objective) for Composite relative to BlackBox-GP, both in percentage (\%) (mean $\pm$ SE over 20 paired runs). Positive values favor Composite. Bold marks unadjusted $p<0.05$ (one-sided paired $t$-tests for improvement).
}
    \begin{tabular}{|l|c|cc|cc|}
        \hline
        & & \multicolumn{2}{c|}{Single objective}
          & \multicolumn{2}{c|}{Multi-objective} \\
        Outer mapping $g$ & Bijective? & Final-RG reduction (\%) & $p$ & Final-HV gain (\%) & $p$ \\
        \hline
                Identity
        & Yes
        & $+18.1 \pm 29.2$ & $0.29$
        & $+0.2 \pm 0.9$ & $0.40$ \\

        Cubic $h^3$
        & Yes
        & $+17.1 \pm 21.1$ & $0.24$
        & $+0.9 \pm 1.6$ & $0.29$ \\
        \hline
        Square $h^2$
        & No
        & $\mathbf{+99.8 \pm 0.1}$ & $\mathbf{<0.01}$
        & $\mathbf{+2.9 \pm 1.0}$ & $\mathbf{<0.01}$ \\

        Shifted square $(h-3)^2$
        & No
        & $\mathbf{+99.6 \pm 0.5}$ & $\mathbf{<0.01}$
        & $+1.0 \pm 0.8$ & $0.12$ \\

        Three-root cubic
        & No
        & $\mathbf{+45.3 \pm 16.4}$ & $\mathbf{0.03}$
        & $\mathbf{+2.5 \pm 1.2}$ & $\mathbf{0.03}$ \\

        Cosine $\cos h$
        & No
        & $\mathbf{+96.3 \pm 4.1}$ & $\mathbf{<0.01}$
        & $-1.5 \pm 0.9$ & $0.95$ \\
        \hline
    \end{tabular}

    \label{tab:controlled-results}
\end{table}

\paragraph{Controlled Synthetic Experiments:} We fix scalar intermediate functions $h:[0,1]^6\to\mathbb{R}$
and vary the known outer mapping $g$ (See Appendix~\ref{app:h_construction} for $h$ construction).
We compare BlackBox-GP and Composite methods, each of which fits its own GP models
and share the initial design, acquisition rule, and
evaluation budget. Single-objective experiments use LogEI~\cite{jones1998efficient,ament2023unexpected} and we report mean percentage of final $\log_{10}$ simple regret (Final-RG) reduction for Composite relative to BlackBox-GP with standard error (SE).
For multi-objective, we use
$
f_1(\mathbf{x})=g(h_1(\mathbf{x})),$ and $f_2(\mathbf{x})=h_2(\mathbf{x}),
$
where $h_1$ and $h_2$ are independent scalar intermediates, modeled with independent GPs.
The $f_2$ identity mapping isolates composition in $f_1$.
We use LogEHVI~\cite{daulton2020differentiable,ament2023unexpected}
and report mean percentage of final hypervolume (Final-HV) gain against a shared reference point for Composite relative to BlackBox-GP with SE. Higher metrics favor Composite. Single-objective starts with 5 initial points and two-objective starts with 10; both then sequentially evaluate 40 additional points based on their acquisition rules. Appendix~\ref{app:metrics} details comparison metrics.
Four function draws and five initial designs per draw yield 20 paired runs.

Table~\ref{tab:controlled-results}  shows significant single-objective gains for all four selected non-bijective mappings, while for multi-objective cases, only square and three-root cubic yield significant hypervolume gains. Neither selected bijective mapping shows a significant gain in either setting. Square gives the largest gain of the six in both settings, but the two scales are not comparable: its single-objective gain is a 99.8\% reduction in final regret, against a hypervolume gain under 3\%. Hypervolume optimization requires discovering a set of trade-offs rather than a single optimum, which may limit Composite's gains under a fixed evaluation budget. Larger-budget experiments could test this explanation. 

\begin{table}[t]
    \centering
    \caption{Scientific-design experiments. Final hypervolume gains (\%, mean $\pm$ SE) for Composite relative to BlackBox-GP over 10 paired trials per benchmark (80 for Synthetic~1). Positive values favor Composite. Bold marks unadjusted $p<0.05$ (one-sided paired $t$-tests for improvement). \textit{Affine} marks affine outer mappings.
    }
    \label{tab:benchmark-summary}

    \footnotesize
    \setlength{\tabcolsep}{3pt}
    \resizebox{\linewidth}{!}{%
    \begin{tabular}{
        l l r r
        @{\hspace{8pt}\vrule\hspace{8pt}}
        l l r r
    }
    \toprule
    \multicolumn{4}{c}{Low-dimensional} &
    \multicolumn{4}{c}{High-dimensional} \\
    \cmidrule(lr){1-4}
    \cmidrule(lr){5-8}
    Benchmark & Method & Final-HV gain (\%) & $p$ &
    Benchmark & Method & Final-HV gain (\%) & $p$ \\
    \midrule

    Color formulation~\cite{chaoucha2022color}
    & LogEHVI & {\boldmath $+13.8 \pm 1.6$} & {\boldmath $<0.01$} &
    Synthetic 2~\cite{simulationlib}
    & Sp-STCH & {\boldmath $+3.6 \pm 1.8$} & {\boldmath $0.038$} \\
    & STCH & {\boldmath $+8.7 \pm 1.7$} & {\boldmath $<0.01$} &
    & MORBO & {\boldmath $+5.3 \pm 2.6$} & {\boldmath $0.036$} \\
    \addlinespace

    Synthetic 1~\cite{deb2002scalable}
    & LogEHVI & {\boldmath $+12.7 \pm 0.6$} & {\boldmath $<0.01$} &
    Lake management~\cite{ward2015tipping}
    & Sp-STCH & {\boldmath $+8.7 \pm 0.7$} & {\boldmath $<0.01$} \\
    & STCH & {\boldmath $+7.5 \pm 0.4$} & {\boldmath $<0.01$} &
    & MORBO & {\boldmath $+4.1 \pm 1.5$} & {\boldmath $0.012$} \\
    \addlinespace

    Truss design ({\it Affine})~\cite{tanabe2020realworld}
    & LogEHVI & {\boldmath $+0.8 \pm 0.1$} & {\boldmath $<0.01$} &
    Power flow ({\it Affine})~\cite{kumar2021benchmark}
    & Sp-STCH & $+0.2 \pm 0.2$ & $0.17$ \\
    & STCH & {\boldmath $+0.7 \pm 0.3$} & {\boldmath $0.022$} &
    & MORBO & {\boldmath $+1.1 \pm 0.2$} & {\boldmath $<0.01$} \\
    \addlinespace

    Nanoparticle design~\cite{kim2022deep}
    & LogEHVI & $+0.2 \pm 0.2$ & $0.17$ &
    & & & \\
    & STCH & $+0.3 \pm 0.4$ & $0.24$ &
    & & & \\
    \addlinespace

    Reaction optimization~\cite{felton2021summit}
    & LogEHVI & $+0.2 \pm 0.2$ & $0.17$ &
    & & & \\
    & STCH & $-0.4 \pm 0.9$ & $0.67$ &
    & & & \\

    \bottomrule
    \end{tabular}%
    }
\end{table}
\paragraph{Scientific-Design Experiments:} We compare BlackBox-GP and Composite versions of four MOBO methods on eight benchmarks as listed in Table~\ref{tab:benchmark-summary} (see Appendix~\ref{app:benchmarks} for details).
All have at least one non-bijective outer mapping; only truss design and power flow have affine outer mappings. Low-dimensional problems use LogEHVI~\cite{daulton2020differentiable,ament2023unexpected} and STCH~\cite{lin2024smooth,pires2026composite} methods with 5 initial and 40 adaptive evaluations; high-dimensional problems use STCH with a spherical kernel (sp-STCH)~\cite{doumont2026linear} and MORBO~\cite{daulton2022morbo} with 20 initial and 100 adaptive evaluations. Appendices~\ref{app:methods} and~\ref{app:configuration} detail methods and implementation. Composite models intermediates with independent GPs. Two methods share the initial design, acquisition rule, reference point, and budget. We report mean percentage gains in Final-HV $\pm$ SE over 10 different trials for all problems, except Synthetic 1, which uses 80 trials.

Table~\ref{tab:benchmark-summary} shows that Composite improved mean final hypervolume in 15 of 16 comparisons, with the largest gains on color formulation (8.7--13.8\%) and Synthetic~1 (7.5--12.7\%). 
Benefits extend to high-dimensional problems:
Synthetic~2 ($3.6$--$5.3\%$) and lake management ($4.1$--$8.7\%$).
Composite provides statistically significant gains on the two affine benchmarks. This might seem to contradict Proposition~\ref{prob:equivalence}: for affine $g$, BlackBox-Induced also places a GP prior on $f$, yet observing intermediates provides no additional predictive information under that shared prior. However, the GP priors used by BlackBox-Induced and BlackBox-GP need not coincide. In our implementation, each intermediate has its own fitted kernel, so the proposition's separable-covariance condition generally fails. Combining the kernels fitted to individual intermediates induces a richer kernel family over $f$ than BlackBox-GP allows, potentially explaining these gains.
Nanoparticle design and reaction optimization show small gains despite non-affine, non-bijective mappings, so these properties alone do not explain performance. Both have correlated intermediates that we model with independent GPs; modeling these dependencies may improve performance. Synthetic~2 also shows modest gains despite non-affine mappings. Limited data in this high-dimensional problem may hinder both approaches. Independently varying input dimension and evaluation budget could clarify their effects.

\section{Conclusion}
We investigate when modeling intermediate responses improves MOBO through theory and experiments. Under a shared prior, intermediate observations add no predictive information when the outer mapping is bijective or affine with separable covariance. In the controlled experiments, large single-objective regret reductions did not translate into comparably large hypervolume gains. 
In scientific-design benchmarks, Composite improved mean final hypervolume in almost all cases, with the largest gains on problems with non-affine, non-bijective outer mappings, although these properties alone do not ensure improvement. Further work should examine how modeling choices such as kernels and dependencies among responses, and input dimension relative to evaluation budget jointly determine these gains.


\section*{Acknowledgement}
We would like to thank the anonymous reviewers for their constructive feedback and suggestions. This material is based on work supported by the AI Research Institutes program supported by the NSF and Intel Corporation under NSF award DMR-2433348. PB gratefully acknowledges financial support from the Development and Promotion of Science and Technology Talents Project (DPST) scholarship program of the Institute for the Promotion of Teaching Science and Technology (IPST), Ministry of Education, Thailand.
\bibliographystyle{unsrt}
\bibliography{references}

\counterwithin{figure}{section}
\counterwithin{table}{section}
\counterwithin{equation}{section}
\newpage
\appendix
{\Large\textbf{Appendix}}
\section{Induced Prior and Inference for BlackBox-Induced}
\label{app:induced}

At fixed evaluation locations $\mathbf{x}_1,\ldots,\mathbf{x}_n$, let $
H=(\mathbf{h}(\mathbf{x}_1),\ldots,\mathbf{h}(\mathbf{x}_n)),$ and $
F=(f(\mathbf{x}_1),\ldots,f(\mathbf{x}_n)).$ BlackBox-Induced and Composite share the vector-valued GP prior
$\mathbf{h} \sim \mathcal{GP}(\mathbf{m}_h, K_h)$, where
$\mathbf{m}_h(\mathbf{x}) = \mathbb{E}[\mathbf{h}(\mathbf{x})]$
is the mean vector and
$K_h(\mathbf{x}, \mathbf{x}') =
\operatorname{Cov}(\mathbf{h}(\mathbf{x}), \mathbf{h}(\mathbf{x}'))$
is the matrix-valued covariance kernel.
This prior induces a prior on $f = g \circ \mathbf{h}$.
BlackBox-Induced conditions only on the objective observations $F$. Its posterior conditions on
$g(\mathbf{h}(\mathbf{x}_j))=F_j$. For bijective $g$, this reduces to standard
GP conditioning on $\mathbf{h}(\mathbf{x}_j)=g^{-1}(F_j)$. For general
non-bijective $g$, inference integrates over intermediate
responses consistent with $F$ and need not have a closed form.
One numerical approach is to repeatedly draw the latent intermediate
responses $H$ from their conditional distribution given $F$ under the
shared GP prior. For each draw of $H$, condition the GP on those
responses, sample $\mathbf{h}(\mathbf{x}_*)$ at a new input
$\mathbf{x}_*$, and apply $g$ to obtain a posterior predictive sample
of $f(\mathbf{x}_*)$. Averaging these samples estimates the posterior
mean $\mathbb{E}[f(\mathbf{x}_*) \mid F]$.

For affine $g(\mathbf{z})=\mathbf{a}^\top \mathbf{z}+b$, the induced prior of $f$ is a GP with
mean $\mathbf{a}^\top \mathbf{m}_h(\mathbf{x})+b$ and kernel $\mathbf{a}^\top K_h(\mathbf{x},\mathbf{x}')\mathbf{a}$.
Inference therefore uses standard GP conditioning directly
on $F$. If the intermediate responses are modeled as independent GPs, then
$K_h(\mathbf{x},\mathbf{x}')$ is diagonal. Its $j$th diagonal entry,
$k_j(\mathbf{x},\mathbf{x}') = [K_h(\mathbf{x},\mathbf{x}')]_{jj}$,
is the covariance kernel of the $j$th intermediate response $h_j$.
The induced kernel of $f$ therefore simplifies to
\[
    \mathbf{a}^\top K_h(\mathbf{x},\mathbf{x}')\mathbf{a}
    = \sum_j a_j^2 k_j(\mathbf{x},\mathbf{x}').
\] A sum of Mat\'ern or RBF kernels
with different lengthscales is generally not a single kernel
of the same family. Thus, separately fitted BlackBox-GP and
BlackBox-Induced priors need not coincide.
\section{Proof of Proposition~\ref{prob:equivalence}}
\label{app:proof-posterior-equivalence}
Throughout, the evaluation locations $\mathbf{x}_1,\ldots,\mathbf{x}_n$ are fixed,
and $H$ and $F$ contain noiseless observations. Both conditional
distributions are computed under a single prior on $h$, with
$f=g\circ h$; in particular, the prior on $f$ is the one induced
by this construction.

For the bijective case, we additionally assume that $g$ is
measurable and bijective on a measurable set containing the
intermediate responses with probability one, and that its inverse
on the corresponding image is measurable. For the affine case,
the mean function and covariance parameters of the GP prior on
$h$ are fixed and known.
\begin{proof}[Proof for the bijective case]
Since $g$ is measurable, the objective observations
\[
F=\bigl(g(\mathbf{h}(\mathbf{x}_1)),\ldots,
        g(\mathbf{h}(\mathbf{x}_n))\bigr)
\]
are a measurable function of $H$. Conversely, since $g$ is bijective
with measurable inverse,
\[
H=\bigl(g^{-1}(f(\mathbf{x}_1)),\ldots,
        g^{-1}(f(\mathbf{x}_n))\bigr)
\]
is a measurable function of $F$. Thus, each dataset can be recovered
from the other, and neither provides additional information when
the other is known.

Consequently, under the same prior on $h$, for every
$x\in\mathcal{X}$ and $t\in\mathbb{R}$,
\[
\begin{aligned}
\mathbb{P}\!\left(f(\mathbf{x})\le t\mid H\right)
&=\mathbb{P}\!\left(f(\mathbf{x})\le t\mid H,F\right)\\
&=\mathbb{P}\!\left(f(\mathbf{x})\le t\mid F\right)
\quad\text{almost surely}.
\end{aligned}
\]
Hence the conditional CDFs coincide.
\end{proof}
\begin{proof}[Proof for the affine case]
Suppose $g(\mathbf z)=\mathbf a^\top\mathbf z+b$ and
$\mathbf h$ has a GP prior with mean $\mathbf m_{\mathbf{h}}(\mathbf{x})$ and separable
covariance
\[
\operatorname{Cov}\!\left(
 \mathbf{h}(\mathbf{x}),\mathbf{h}(\mathbf{x}')
\right)
=k_\mathbf{h}(\mathbf{x},\mathbf{x}')B,
\]
where $k$ is a scalar covariance kernel and $B$ is a fixed
positive semidefinite matrix. Then $f$ is a GP with mean
$m_f(\mathbf{x})=\mathbf a^\top\mathbf m_\mathbf{h}(\mathbf{x})+b$
and covariance
\[
k_f(\mathbf{x},\mathbf{x}')
=c\,k_\mathbf{h}(\mathbf{x},\mathbf{x}'),
\qquad c=\mathbf a^\top B\mathbf a.
\]
If $c=0$, $f$ is deterministic under the prior, so the result
holds immediately. Assume henceforth that $c>0$.

Fix a prediction location $x$ and define
\[
(K_\mathbf{h})_{ij}=k_\mathbf{h}(\mathbf{x}_i,\mathbf{x}_j),
\qquad
\mathbf k_\mathbf{h}=
\bigl(k_\mathbf{h}(\mathbf{x}_1,\mathbf{x}),\ldots,
      k_\mathbf{h}(\mathbf{x}_n,\mathbf{x})\bigr)^\top.
\]
Write $K_\mathbf{h}^\dagger$ for the Moore--Penrose pseudoinverse of $K_\mathbf{h}$,
which equals $K_\mathbf{h}^{-1}$ when $K_\mathbf{h}$ is nonsingular.

Conditioning the vector-valued GP on $H$ gives
\[
\begin{aligned}
\mathbb E[\mathbf{h}(\mathbf{x})\mid H]
&=\mathbf m_\mathbf{h}(\mathbf{x})
+\sum_{i=1}^{n}(K_\mathbf{h}^\dagger\mathbf k_\mathbf{h})_i
 \bigl(\mathbf{h}(\mathbf{x}_i)-\mathbf m_\mathbf{h}(\mathbf{x}_i)\bigr),\\
\operatorname{Cov}(\mathbf{h}(\mathbf{x})\mid H)
&=\bigl(k_\mathbf{h}(\mathbf{x},\mathbf{x})
-\mathbf k_\mathbf{h}^\top K_\mathbf{h}^\dagger\mathbf k_\mathbf{h}\bigr)B.
\end{aligned}
\]
Applying $g$ therefore yields a Gaussian conditional distribution
for $f(\mathbf{x})$ with mean and variance
\[
\begin{aligned}
\mu_H(\mathbf{x})
&=m_f(\mathbf{x})
+\mathbf k_\mathbf{h}^\top K_\mathbf{h}^\dagger
 \bigl(F-\mathbf m_f(\mathbf{X})\bigr),\\
v_H(\mathbf{x})
&=c\bigl(k_\mathbf{h}(\mathbf{x},\mathbf{x})
-\mathbf k_\mathbf{h}^\top K_\mathbf{h}^\dagger\mathbf k_\mathbf{h}\bigr),
\end{aligned}
\]
where
$\mathbf m_f(\mathbf{X})=(m_f(\mathbf{x}_1),\ldots,m_f(\mathbf{x}_n))^\top$.

Alternatively, conditioning the scalar GP $f$ directly on $F$
gives
\[
\begin{aligned}
\mu_F(\mathbf{x})
&=m_f(\mathbf{x})
+(c\mathbf k_\mathbf{h})^\top(cK_\mathbf{h})^\dagger
 \bigl(F-\mathbf m_f(\mathbf{X})\bigr)
=\mu_H(\mathbf{x}),\\
v_F(\mathbf{x})
&=c\,k_\mathbf{h}(\mathbf{x},\mathbf{x})
-(c\mathbf k_\mathbf{h})^\top(cK_\mathbf{h})^\dagger(c\mathbf k_\mathbf{h})
=v_H(\mathbf{x}),
\end{aligned}
\]
using $(cK_\mathbf{h})^\dagger=c^{-1}K_\mathbf{h}^\dagger$.

Both conditional distributions are Gaussian with identical means
and variances. Hence, for every $t\in\mathbb R$,
\[
\mathbb P\!\left(f(\mathbf{x})\le t\mid H\right)
=
\mathbb P\!\left(f(\mathbf{x})\le t\mid F\right)
\quad\text{almost surely}.
\]
\end{proof}
\section{Details of the construction of $h$ function in controlled experiments}
\label{app:h_construction}
In the controlled experiments, for each $h$, we randomly generate $h$ using 1024 random Fourier features approximating a Matern-5/2 GP with lengthscale 0.5. For each problem, this $h$ function is randomly generated four times.
\section{Comparison metrics}
\label{app:metrics}
\subsection{Single-objective metric}
Let $f^{\star}=\min_\mathbf{x} f(\mathbf{x})$ be the optimum, computed numerically for every
controlled mapping, and let
\[
r_T^{v,s}=\Bigl(\min_{1\leq t\leq T}f(\mathbf{x}_t^{v,s})\Bigr)-f^{\star}
\]
be the simple regret of variant $v\in\{\mathrm{BlackBox\text{-}GP},\mathrm{Composite}\}$
in paired trial $s$ within evaluation budget $T$. Since we minimize, smaller
values indicate better performance. Regret varies over many orders of
magnitude, so we average it on the $\log_{10}$ scale:
\[
\overline{r}_T^{v}=10^{\,m_T^{v}},
\qquad
m_T^{v}=\frac{1}{S}\sum_{s=1}^{S}\log_{10}r_T^{v,s},
\]
where $m_T^{v}$ is the BlackBox-GP or Composite column of
Table~\ref{tab:so-full}. We measure final performance by
\begin{equation}
\text{Final-RG reduction}(\%)
=100\,
\frac{\overline{r}_T^{\mathrm{BlackBox\text{-}GP}}
      -\overline{r}_T^{\mathrm{Composite}}}
{\overline{r}_T^{\mathrm{BlackBox\text{-}GP}}}.
\label{eqn:final_gain}
\end{equation}
Positive reductions favor Composite.
\subsection{Multi-objective metric}
Let $\mathcal{P}\subset\mathbb{R}^N$ be the set of nondominated
objective vectors observed so far for a minimization problem. A vector
$\mathbf{y}$ dominates $\mathbf{y}'$ if $y_i\leq y_i'$ for every $i$
and the inequality is strict for at least one objective. Given a fixed
reference point $\mathbf{r}$ worse than every vector in $\mathcal{P}$,
the dominated hypervolume (HV) is
\begin{equation}
\operatorname{HV}(\mathcal{P};\mathbf{r})
=\lambda_N\!\left(
\bigcup_{\mathbf{y}\in\mathcal{P}}
\prod_{i=1}^{N}[y_i,r_i]
\right),
\label{eq:hypervolume}
\end{equation}
where $\lambda_N$ denotes $N$-dimensional volume. Larger HV indicates
better objective values or broader coverage of favorable tradeoffs.
When an observed vector is not componentwise better than the reference
point, it contributes no volume. We use the same fixed reference point
for all methods and trials in a given setting.
Let $\mathrm{HV}^{v,s}_t$ denote the dominated hypervolume after $t$ total
evaluations for variant $v\in\{\mathrm{BlackBox\text{-}GP},\mathrm{Composite}\}$ in
paired trial $s$. The mean HV across $S$ trials is
\[
\overline{\mathrm{HV}}^v_t
=\frac{1}{S}\sum_{s=1}^{S}\mathrm{HV}^{v,s}_t.
\]
We measure final performance by the relative improvement in mean HV at
the total evaluation budget $T$:
\begin{equation}
\text{Final-HV gain}(\%)
=100\frac{\overline{\mathrm{HV}}^{\mathrm{Composite}}_T-
\overline{\mathrm{HV}}^{\mathrm{BlackBox\text{-}GP}}_T}
{\overline{\mathrm{HV}}^{\mathrm{BlackBox\text{-}GP}}_T}.
\label{eq:final-hv-gain}
\end{equation}
\subsection{Significance}
The two arms share the function draw, the initial design, the acquisition rule
and the evaluation budget, so runs are paired. All reported $p$-values are
one-sided paired $t$-tests for a positive mean gain, computed on the per-run
differences $\log_{10}r_T^{\mathrm{BlackBox\text{-}GP},s}-\log_{10}r_T^{\mathrm{Composite},s}$
and $\mathrm{HV}_T^{\mathrm{Composite},s}-\mathrm{HV}_T^{\mathrm{BlackBox\text{-}GP},s}$.
All tables report mean $\pm$ one standard error over the paired runs.
Eq.~\eqref{eqn:final_gain} and Eq.~\eqref{eq:final-hv-gain} are functions of
the arm means rather than averages of per-run percentages, so their standard
errors are obtained from the standard error of the mean paired difference by
the delta method.

\section{Additional results for controlled experiments}
\label{app:controlled}

Table~\ref{tab:controlled-results} of the main text reports six representative outer
mappings. This
appendix gives every mapping we considered, under the metrics of
Appendix~\ref{app:metrics}. Table~\ref{tab:so-full} reports results for the single-objective
cases while Table~\ref{tab:mo-full} reports results for multi-objective (two-objective) cases. Both settings use four function
draws and five initial designs per draw, giving $20$ paired runs per problem,
and within each comparison the two methods (BlackBox-GP and Composite) share the initial design, the
acquisition rule and the evaluation budget.

\paragraph{Single objective.} Of the five mappings that are bijective on the
range $h$ attains, none reaches significance; of the eleven that are not, every
one does, with $p\le 0.030$ throughout. Bijectivity therefore separates the two
groups perfectly in this setting.

Percentage reductions near $100\%$ can obscure substantial differences in final regret. For the two-root quadratic and four-root quartic, reductions of $99.96\%$ and $99.53\%$ correspond to approximately $2600$-fold and $210$-fold reductions relative to BlackBox-GP, respectively. Table~\ref{tab:so-full} therefore also reports gains in $\log_{10}$ regret to distinguish improvements of different orders of magnitude. The 100\% ceiling also explains the narrow standard errors on the saturated rows.

\paragraph{Multi-objective (two objectives).} The pattern transfers only partially. None of the
five bijective mappings reaches significance, as in the single-objective case,
but only five of the fourteen non-bijective settings do, and the effect sizes are
percent-level rather than orders of magnitude. The ordering also fails to carry
over: the three- and five-root polynomials give the two weakest single-objective
gains but both yield significant hypervolume gains, including the largest, while the four-root quartic and the
shifted squares at $(h-1)^{2}$ and $(h-2)^{2}$ all exceed $+99.5\%$ in the
single-objective grid yet fall below $+0.5\%$ here. Hypervolume optimization
requires discovering and refining a set of
tradeoffs rather than locating a single optimum, and under a fixed budget that
additional difficulty appears to dilute the benefit of modelling intermediates.

\begin{table}[!htbp]
\centering\footnotesize
\caption{Controlled single-objective experiments, all $16$ outer mappings. Values in BlackBox-GP and Composite columns are the final $\log_{10}$ simple regret from each method; Reduction is their paired difference, so positive favors Composite. Final-RG reduction is that same difference as a percentage, via Eq.~\eqref{eqn:final_gain}. All values are means $\pm$ one standard error over the $20$ paired runs, the percentage column by the delta method, and $p$ is a one-sided paired $t$-test for positive mean reduction. Bold marks significance at the unadjusted $0.05$ level.}
\label{tab:so-full}
\setlength{\tabcolsep}{4.5pt}
\begin{tabular}{l r r r r r}
\toprule
Outer mapping $g$ & BlackBox-GP & Composite & Reduction & Final-RG reduction (\%) & $p$ \\
\midrule
\multicolumn{6}{l}{\emph{Bijective on the range $h$ attains}}\\
Exponential $e^{h}$ & $-2.59 \pm 0.09$ & $-2.74 \pm 0.14$ & $+0.15 \pm 0.12$ & $+28.91 \pm 19.01$ & 0.11 \\
Identity $h$ & $-1.03 \pm 0.18$ & $-1.11 \pm 0.14$ & $+0.09 \pm 0.16$ & $+18.10 \pm 29.24$ & 0.29 \\
Cubic $h^{3}$ & $+0.09 \pm 0.11$ & $+0.01 \pm 0.10$ & $+0.08 \pm 0.11$ & $+17.10 \pm 21.07$ & 0.23 \\
Hyperbolic sine $\sinh h$ & $+0.24 \pm 0.11$ & $+0.16 \pm 0.09$ & $+0.08 \pm 0.10$ & $+17.00 \pm 19.58$ & 0.22 \\
Quintic $h^{5}$ & $+0.75 \pm 0.15$ & $+0.79 \pm 0.08$ & $-0.05 \pm 0.16$ & $-11.62 \pm 40.30$ & 0.62 \\
\midrule
\multicolumn{6}{l}{\emph{Not bijective}}\\
Two-root quadratic & $\mathbf{-4.97 \pm 0.28}$ & $\mathbf{-8.39 \pm 0.30}$ & $\mathbf{+3.42 \pm 0.41}$ & $\mathbf{+99.96 \pm 0.04}$ & \textbf{$<$0.01} \\
Shifted square $(h-1)^{2}$ & $\mathbf{-5.36 \pm 0.24}$ & $\mathbf{-8.22 \pm 0.24}$ & $\mathbf{+2.86 \pm 0.30}$ & $\mathbf{+99.86 \pm 0.09}$ & \textbf{$<$0.01} \\
Square $h^{2}$ & $\mathbf{-5.14 \pm 0.25}$ & $\mathbf{-7.95 \pm 0.26}$ & $\mathbf{+2.81 \pm 0.40}$ & $\mathbf{+99.84 \pm 0.14}$ & \textbf{$<$0.01} \\
Shifted square $(h-2)^{2}$ & $\mathbf{-5.62 \pm 0.22}$ & $\mathbf{-8.40 \pm 0.28}$ & $\mathbf{+2.78 \pm 0.35}$ & $\mathbf{+99.83 \pm 0.13}$ & \textbf{$<$0.01} \\
Shifted square $(h-0.5)^{2}$ & $\mathbf{-5.40 \pm 0.28}$ & $\mathbf{-7.81 \pm 0.14}$ & $\mathbf{+2.41 \pm 0.25}$ & $\mathbf{+99.61 \pm 0.22}$ & \textbf{$<$0.01} \\
Shifted square $(h-3)^{2}$ & $\mathbf{-4.69 \pm 0.36}$ & $\mathbf{-7.05 \pm 0.49}$ & $\mathbf{+2.36 \pm 0.46}$ & $\mathbf{+99.56 \pm 0.46}$ & \textbf{$<$0.01} \\
Four-root quartic & $\mathbf{-4.80 \pm 0.21}$ & $\mathbf{-7.13 \pm 0.17}$ & $\mathbf{+2.33 \pm 0.20}$ & $\mathbf{+99.53 \pm 0.22}$ & \textbf{$<$0.01} \\
Cosine $\cos h$ & $\mathbf{-4.36 \pm 0.41}$ & $\mathbf{-5.79 \pm 0.43}$ & $\mathbf{+1.43 \pm 0.48}$ & $\mathbf{+96.31 \pm 4.06}$ & \textbf{$<$0.01} \\
Cosine $\cos 2h$ & $\mathbf{-4.94 \pm 0.32}$ & $\mathbf{-6.00 \pm 0.20}$ & $\mathbf{+1.06 \pm 0.35}$ & $\mathbf{+91.27 \pm 7.08}$ & \textbf{$<$0.01} \\
Five-root quintic & $\mathbf{+1.69 \pm 0.04}$ & $\mathbf{+1.27 \pm 0.16}$ & $\mathbf{+0.42 \pm 0.14}$ & $\mathbf{+62.36 \pm 12.33}$ & \textbf{$<$0.01} \\
Three-root cubic & $\mathbf{+0.54 \pm 0.09}$ & $\mathbf{+0.28 \pm 0.11}$ & $\mathbf{+0.26 \pm 0.13}$ & $\mathbf{+45.32 \pm 16.44}$ & \textbf{0.03} \\
\bottomrule
\end{tabular}
\end{table}

\begin{table}[!htbp]
\centering\footnotesize
\caption{Controlled two-objective experiments, all $19$ distinct settings of $g_1$. Values in BlackBox-GP and Composite columns are the final hypervolume; Gain is their paired difference, so positive favors Composite. Final-HV gain is that same difference as a percentage, via Eq.~\eqref{eq:final-hv-gain}. All values are means $\pm$ one standard error over the $20$ paired runs, the percentage column by the delta method, and $p$ is a one-sided paired $t$-test for positive mean gain. The BlackBox-GP and Composite columns have the largest standard errors because each mapping pools four function draws whose hypervolumes differ substantially; that spread cancels in the paired Gain. Bold marks significance at the unadjusted $0.05$ level.}
\label{tab:mo-full}
\setlength{\tabcolsep}{4.5pt}
\begin{tabular}{l r r r r r}
\toprule
Outer mapping $g_1$ & BlackBox-GP & Composite & Gain & Final-HV gain (\%) & $p$ \\
\midrule
\multicolumn{6}{l}{\emph{Bijective on the range $h$ attains}}\\
Cubic $h^{3}$ & $71.49 \pm 11.17$ & $72.12 \pm 11.22$ & $+0.63 \pm 1.11$ & $+0.88 \pm 1.56$ & 0.29 \\
Exponential $e^{h}$ & $71.17 \pm 11.22$ & $71.58 \pm 11.09$ & $+0.41 \pm 0.92$ & $+0.58 \pm 1.29$ & 0.33 \\
Quintic $h^{5}$ & $71.95 \pm 11.11$ & $72.21 \pm 10.97$ & $+0.26 \pm 1.44$ & $+0.36 \pm 2.00$ & 0.43 \\
Identity $h$ & $70.93 \pm 11.15$ & $71.10 \pm 11.30$ & $+0.16 \pm 0.65$ & $+0.23 \pm 0.92$ & 0.40 \\
Hyperbolic sine $\sinh h$ & $71.46 \pm 11.25$ & $71.19 \pm 11.21$ & $-0.27 \pm 0.53$ & $-0.38 \pm 0.75$ & 0.69 \\
\midrule
\multicolumn{6}{l}{\emph{Not bijective}}\\
Five-root quintic & $\mathbf{72.13 \pm 11.27}$ & $\mathbf{80.51 \pm 10.04}$ & $\mathbf{+8.38 \pm 3.23}$ & $\mathbf{+11.62 \pm 4.48}$ & \textbf{0.01} \\
Square $h^{2}$ & $\mathbf{69.88 \pm 10.87}$ & $\mathbf{71.91 \pm 11.16}$ & $\mathbf{+2.03 \pm 0.70}$ & $\mathbf{+2.90 \pm 1.01}$ & \textbf{$<$0.01} \\
Shifted square $(h-0.5)^{2}$ & $\mathbf{70.98 \pm 11.03}$ & $\mathbf{72.94 \pm 11.28}$ & $\mathbf{+1.96 \pm 0.80}$ & $\mathbf{+2.76 \pm 1.13}$ & \textbf{0.01} \\
Three-root cubic & $\mathbf{70.72 \pm 11.12}$ & $\mathbf{72.52 \pm 11.04}$ & $\mathbf{+1.80 \pm 0.86}$ & $\mathbf{+2.54 \pm 1.22}$ & \textbf{0.03} \\
Two-root quadratic & $71.56 \pm 11.23$ & $72.88 \pm 11.23$ & $+1.32 \pm 0.81$ & $+1.85 \pm 1.13$ & 0.06 \\
Cosine $\cos 2h$ & $\mathbf{70.54 \pm 11.05}$ & $\mathbf{71.39 \pm 11.08}$ & $\mathbf{+0.85 \pm 0.45}$ & $\mathbf{+1.20 \pm 0.64}$ & \textbf{0.04} \\
Shifted square $(h-3)^{2}$ & $70.79 \pm 11.17$ & $71.50 \pm 11.18$ & $+0.70 \pm 0.58$ & $+1.00 \pm 0.83$ & 0.12 \\
Three-root, spread $0.5$ & $70.78 \pm 10.95$ & $71.31 \pm 10.95$ & $+0.53 \pm 0.57$ & $+0.75 \pm 0.80$ & 0.18 \\
Four-root, spread $0.5$ & $71.15 \pm 11.08$ & $71.63 \pm 11.22$ & $+0.47 \pm 0.55$ & $+0.66 \pm 0.78$ & 0.20 \\
Shifted square $(h-1)^{2}$ & $71.52 \pm 11.29$ & $71.87 \pm 11.19$ & $+0.35 \pm 0.65$ & $+0.49 \pm 0.90$ & 0.30 \\
Shifted square $(h-2)^{2}$ & $71.09 \pm 11.02$ & $71.39 \pm 11.19$ & $+0.30 \pm 0.84$ & $+0.42 \pm 1.18$ & 0.36 \\
Six-root, spread $1.0$ & $71.61 \pm 11.19$ & $71.66 \pm 11.14$ & $+0.05 \pm 1.06$ & $+0.07 \pm 1.48$ & 0.48 \\
Four-root quartic & $71.97 \pm 11.40$ & $71.95 \pm 11.28$ & $-0.01 \pm 0.85$ & $-0.02 \pm 1.18$ & 0.51 \\
Cosine $\cos h$ & $71.86 \pm 11.18$ & $70.81 \pm 11.15$ & $-1.05 \pm 0.63$ & $-1.46 \pm 0.87$ & 0.95 \\
\bottomrule
\end{tabular}
\end{table}

\clearpage

\section{Benchmark definitions}
\label{app:benchmarks}

This section specifies the exact deterministic oracle used in the experiments.
All design vectors $\mathbf{x}$ below are normalized to $[0,1]^d$, and all
final objectives are minimized. Here $d$ is the input dimension, $K$ is the
number of scalar intermediate responses retained by the implementation, and $N$ is the number of final objectives.

\subsection{Multi-illuminant color formulation: $d=6$, $K=6$, $N=2$}

This analytical simulator represents six hypothetical dye loadings in an
opaque textile coating.  It is motivated by multi-objective color formulation
under multiple illuminants \cite{chaoucha2022color}, but is not a measured
dataset and does not reproduce that paper's objective equations.  Wavelengths
are $\lambda\in\{400,410,\ldots,700\}$ nm. Define the Gaussian spectral bump of center $\mu$ and width $\sigma$
\begin{equation}
 B(\lambda;\mu,\sigma)=
 \exp\!\left[-\tfrac12\left(\frac{\lambda-\mu}{\sigma}\right)^2\right].
\end{equation}
The substrate absorption-to-scattering ratio is
$q_0(\lambda)=0.012+0.008B(\lambda;430,75)$. The six fixed dye
increments are
\begin{equation}
\begin{aligned}
q_1(\lambda) &= 1.75B(\lambda;610,34)+0.22B(\lambda;445,52),\\
q_2(\lambda) &= 1.55B(\lambda;535,31)+0.18B(\lambda;650,42),\\
q_3(\lambda) &= 1.65B(\lambda;448,28)+0.16B(\lambda;585,65),\\
q_4(\lambda) &= 1.20B(\lambda;495,48)+0.38B(\lambda;632,27),\\
q_5(\lambda) &= 1.35B(\lambda;575,41)+0.24B(\lambda;420,24),\\
q_6(\lambda) &= 0.55+0.22B(\lambda;510,115).
\end{aligned}
\end{equation}
For design $\mathbf{x}$, the dye concentrations and total
absorption-to-scattering ratio are
\begin{equation}
c_m(\mathbf{x})=0.85x_m,\qquad
q(\lambda;\mathbf{x})
=
q_0(\lambda)+\sum_{m=1}^{6}c_m(\mathbf{x})q_m(\lambda).
\end{equation}
The opaque Kubelka--Munk inverse \cite{kubelka1931optik} gives
\begin{equation}
R(\lambda;\mathbf{x})
=
1+q(\lambda;\mathbf{x})
-\sqrt{q(\lambda;\mathbf{x})^2+2q(\lambda;\mathbf{x})}.
\label{eq:km}
\end{equation}

Let
$\overline{\mathbf{c}}(\lambda)
=(\bar{x}(\lambda),\bar{y}(\lambda),\bar{z}(\lambda))$
denote the analytic CIE 1931 color-matching functions of Wyman et al.\
\cite{wyman2013simple}, and write
$\Lambda_c=\{400,410,\ldots,700\}$ nm. For
$\lambda_{\mathrm m}=10^{-9}\lambda$ meters, define the normalized blackbody
spectrum
\begin{equation}
\widetilde I_T(\lambda)
=
\frac{1}{
\lambda_{\mathrm m}^{5}
\left[
\exp\!\left(\frac{c_2}{\lambda_{\mathrm m}T}\right)-1
\right]},
\qquad
I_T(\lambda)
=
\frac{\widetilde I_T(\lambda)}
{\max_{\lambda'\in\Lambda_c}\widetilde I_T(\lambda')},
\end{equation}
where $c_2=1.438776877\times10^{-2}\ {\rm m\,K}$. For illuminant
$j\in\{D,A\}$, with $T_D=6504$ K and $T_A=2856$ K, define
\begin{equation}
\kappa_j
=
\left[
\sum_{\lambda\in\Lambda_c}
I_{T_j}(\lambda)\bar{y}(\lambda)
\right]^{-1}.
\end{equation}
The tristimulus coordinates and perfect-reflector white point are
\begin{align}
\bigl(X_j(\mathbf{x}),Y_j(\mathbf{x}),Z_j(\mathbf{x})\bigr)
&=
\kappa_j
\sum_{\lambda\in\Lambda_c}
R(\lambda;\mathbf{x})I_{T_j}(\lambda)
\overline{\mathbf{c}}(\lambda),\\
\bigl(X_{n,j},Y_{n,j},Z_{n,j}\bigr)
&=
\kappa_j
\sum_{\lambda\in\Lambda_c}
I_{T_j}(\lambda)\overline{\mathbf{c}}(\lambda).
\end{align}
The corresponding CIE Lab coordinates are
\begin{align}
L_j^*(\mathbf{x})
&=
116\phi\!\left(\frac{Y_j(\mathbf{x})}{Y_{n,j}}\right)-16,\\
a_j^*(\mathbf{x})
&=
500\left[
\phi\!\left(\frac{X_j(\mathbf{x})}{X_{n,j}}\right)
-\phi\!\left(\frac{Y_j(\mathbf{x})}{Y_{n,j}}\right)
\right],\\
b_j^*(\mathbf{x})
&=
200\left[
\phi\!\left(\frac{Y_j(\mathbf{x})}{Y_{n,j}}\right)
-\phi\!\left(\frac{Z_j(\mathbf{x})}{Z_{n,j}}\right)
\right],
\end{align}
where $\delta=6/29$ and
\begin{equation}
\phi(t)=
\begin{cases}
t^{1/3}, & t>\delta^3,\\
t/(3\delta^2)+4/29, & t\leq\delta^3.
\end{cases}
\end{equation} 
The objective-specific intermediate responses are
\begin{equation}
\begin{aligned}
\mathbf{h}_1(\mathbf{x})
&=
\bigl(
L_D^*(\mathbf{x}),
a_D^*(\mathbf{x}),
b_D^*(\mathbf{x})
\bigr),\\
\mathbf{h}_2(\mathbf{x})
&=
\bigl(
L_D^*(\mathbf{x}),
a_D^*(\mathbf{x}),
b_D^*(\mathbf{x}),
L_A^*(\mathbf{x}),
a_A^*(\mathbf{x}),
b_A^*(\mathbf{x})
\bigr).
\end{aligned}
\end{equation}
These contain six distinct retained scalar responses because the daylight
coordinates are shared.

The unattainable target pigment has
\begin{equation}
q_\star(\lambda)
=
q_0(\lambda)
+0.92B(\lambda;482,19)
+0.74B(\lambda;622,25)
+0.18B(\lambda;550,72).
\end{equation}
The target ratio $q_\star(\lambda)$ is converted to reflectance using
Eq.~\eqref{eq:km}, then to illuminant-specific XYZ and CIE Lab coordinates
using the equations above. The resulting target vectors are
$\mathbf{t}_j=(L_{\star,j}^*,a_{\star,j}^*,b_{\star,j}^*)$ for
$j\in\{D,A\}$.
For a generic argument $\mathbf{y}$, define
\begin{equation}
  g_1(\mathbf{y})=\frac{\lVert \mathbf{y}_{1:3}-\mathbf{t}_D\rVert_2}{60},
  \qquad
  g_2(\mathbf{y})=
  \frac{\lVert(\mathbf{y}_{4:6}-\mathbf{t}_A)
  -(\mathbf{y}_{1:3}-\mathbf{t}_D)\rVert_2}{18}.
  \label{eq:color-objectives}
\end{equation}
The objectives are $f_i(\mathbf{x})=g_i(\mathbf{h}_i(\mathbf{x}))$.
Thus $f_1$ is daylight color error and $f_2$ is a metameric-mismatch proxy: it
penalizes how the sample-minus-target Lab error changes between illuminants.
The nonlinear norms erase the direction of the six-dimensional color errors.
The hypervolume reference is $(1.05,1.05)$.
\subsection{DTLZ2 (Synthetic 1): $d=6$, $K=4$, $N=2$}

The synthetic two-objective DTLZ2 problem \cite{deb2002scalable} is
\begin{equation}
 q(\mathbf{x})=\sum_{j=2}^{6}(x_j-0.5)^2,\qquad
 \theta(\mathbf{x})=\frac{\pi x_1}{2},
\end{equation}
\begin{equation}
 \mathbf{h}_1(\mathbf{x})=\big(q(\mathbf{x}),\cos\theta(\mathbf{x})\big),\qquad
 \mathbf{h}_2(\mathbf{x})=\big(q(\mathbf{x}),\sin\theta(\mathbf{x})\big),
 \label{eq:dtlz-intermediates}
\end{equation}
\begin{equation}
 g_1(\mathbf{y})=(1+y_1)y_2,\qquad
 g_2(\mathbf{y})=(1+y_1)y_2,\qquad
 f_i(\mathbf{x})=g_i(\mathbf{h}_i(\mathbf{x})).
 \label{eq:dtlz}
\end{equation}
The repeated $q$ deliberately gives each objective an independently modeled
component group, avoiding implicit multi-task transfer.  The Pareto set has
$x_j=0.5$ for $j=2,\ldots,6$ and the front satisfies
$f_1^2+f_2^2=1$, $f_1,f_2\ge0$.  For reference point $(2.5,2.5)$ the exact
maximum hypervolume is $2.5^2-\pi/4$.
\subsection{Four-bar truss RE21: $d=4$, $K=8$, $N=2$}

Four bars are joined into a frame carrying a load at one joint, and the design chooses how thick each bar is. Thicker bars sag less but use more material, giving the problem a tradeoff. This problem follows the RE21 formulation in the real-world multi-objective suite \cite{tanabe2020realworld}.  The four
cross-sectional areas are
\begin{equation}
 \mathbf{a}(\mathbf{x})=\mathbf{a}^{\min}+\mathbf{x}\odot
 (\mathbf{a}^{\max}-\mathbf{a}^{\min}),\quad
 \mathbf{a}^{\min}=(1,\sqrt2,\sqrt2,1),\quad
 \mathbf{a}^{\max}=(3,3,3,3)
\end{equation}
in cm$^2$, where each entry of $\mathbf{x}$ sets that bar's area as a fraction of the way between its minimum and maximum. The lower bounds are the thinnest each bar can be without exceeding the material's stress limit
of $10$ kN/cm$^2$ under the load $F=10$ kN, so every design in the box is
already safe. Each bar is $L=200$ cm long and the material's stiffness (elastic modulus) is $E=2\times10^5$ kN/cm$^2$. The intermediate responses are
\begin{equation}
\begin{aligned}
\mathbf{h}_1(\mathbf{x})
&=
\left(
2a_1(\mathbf{x}),
\sqrt{2}\,a_2(\mathbf{x}),
\sqrt{a_3(\mathbf{x})},
a_4(\mathbf{x})
\right),\\
\mathbf{h}_2(\mathbf{x})
&=
\left(
\frac{2}{a_1(\mathbf{x})},
\frac{2\sqrt{2}}{a_2(\mathbf{x})},
-\frac{2\sqrt{2}}{a_3(\mathbf{x})},
\frac{2}{a_4(\mathbf{x})}
\right).
\end{aligned}
\end{equation}
Each entry corresponds to a bar's contribution to total material in $\mathbf{h}_1$, and to the sag at the loaded joint in $\mathbf{h}_2$. Raw volume and loaded-joint displacement are
\begin{equation}
 V(\mathbf{y})=L\sum_{j=1}^4y_j,
 \qquad
 D(\mathbf{y})=\frac{FL}{E}\sum_{j=1}^4y_j.
\end{equation}
They are affinely normalized using exact extrema over the box:
\begin{align}
V_{\min}&=1237.8414,& V_{\max}&=2994.9383,\\
D_{\min}&=0.00276142,& D_{\max}&=0.05057191,\\
g_1(\mathbf{y})&=\frac{V(\mathbf{y})-V_{\min}}{V_{\max}-V_{\min}},&
g_2(\mathbf{y})&=\frac{D(\mathbf{y})-D_{\min}}{D_{\max}-D_{\min}}.
\end{align}
The final objectives are $f_i(\mathbf{x})=g_i(\mathbf{h}_i(\mathbf{x}))$.
The reference is $(1.05,1.05)$.  Unlike the color and DTLZ2 reductions, the
outer map is essentially a signed sum plus affine scaling, consistent with the small, though significant, gains in Table~\ref{tab:benchmark-summary}.
\subsection{Latent Langermann--Ackley (Synthetic 2): $d=600$, $K=12$, $N=2$}

Let $A\in\R^{6\times600}$ have fixed orthonormal rows generated once with seed
18,271, and define the dense latent vector
\begin{equation}
 \mathbf{z}(\mathbf{x})=8A(\mathbf{x}-\tfrac12\mathbf1).
\end{equation}
The intermediate responses are objective-specific copies,
$\mathbf{h}_1(\mathbf{x})=\mathbf{h}_2(\mathbf{x})=\mathbf{z}(\mathbf{x})\in\R^6$.
Thus the implementation retains $K=12$ scalar responses, which prevents its independent GP implementation from
sharing the six latent observations across objectives.  For centers
\begin{equation}
C=\begin{bmatrix}
1.6&-0.9&0.7&1.2&-1.4&0.5\\
-1.3&1.5&-0.8&0.6&1.0&-1.1\\
0.8&1.1&1.4&-1.5&-0.5&0.9\\
-0.7&-1.6&1.0&1.4&0.8&-0.6\\
1.2&0.6&-1.5&-0.9&1.3&1.0
\end{bmatrix},\quad \mathbf{c}=(1,2,5,2,3),
\end{equation}
define $r_\ell(\mathbf{y})=\lVert \mathbf{y}-C_\ell\rVert_2^2$ and
$C_\Sigma=\sum_{\ell=1}^{5}c_\ell=13$. The two known mappings are
\begin{align}
g_1(\mathbf{y})&=\frac{C_\Sigma-
 \sum_{\ell=1}^{5}c_\ell\exp[-r_\ell(\mathbf{y})/\pi]
 \cos[\pi r_\ell(\mathbf{y})]}{2C_\Sigma},
 \label{eq:langermann}\\
g_2(\mathbf{y})&=\frac{-20\exp[-0.2\sqrt{\frac16\sum_j y_j^2}]
 -\exp[\frac16\sum_j\cos(2\pi y_j)]+20+e}{20+e-e^{-1}}.
 \label{eq:ackley}
\end{align}
The objectives are $f_i(\mathbf{x})=g_i(\mathbf{h}_i(\mathbf{x}))$.
$f_1$ is a bounded Langermann-style landscape and $f_2$ is normalized Ackley.
Their optima conflict: Ackley is minimized at $\mathbf{z}=\mathbf{0}$, whereas the Langermann
centers are displaced.  The reference point is $(1.1,1.1)$. 
\subsection{Shallow-lake management: $d=100$, $K=21$, $N=2$}
A town releases phosphorus into a nearby lake each year for 100 years. Releasing more is economically useful, but phosphorus builds up in the lake, and past a threshold the lake flips into a polluted state that is very hard to reverse. The design chooses the 100 yearly release amounts.
Each $x_t$ in $\mathbf{x}$ controls annual phosphorus release
$u_t=0.1x_t$ over 100 years.  The lake is simulated under 100 fixed
common-random-number inflow scenarios ($s$). For scenario $s$, writing $P_t^{(s)}$ for the phosphorus in
the lake water at year $t$, we have  $P_0^{(s)}=0$ and
for $t=1,\ldots,99$,
\begin{equation}
P_t^{(s)}=0.58P_{t-1}^{(s)}+
\frac{(P_{t-1}^{(s)})^2}{1+(P_{t-1}^{(s)})^2}
+u_{t-1}+\xi_{t-1}^{(s)},
\label{eq:lake-dynamics}
\end{equation}
where
$\xi_t^{(s)}\sim\operatorname{LogNormal}(-3.52,0.105^2)$ was drawn once with
seed 20,240,820. The first term is the carryover of phosphorus from last year, the third term is the amount the town chose to release last year, and the fourth is uncontrolled inflow from nature. The second term accounts for phosphorus that settled into the lake bed and later returns to the water. Below a critical level the lake clears itself faster than this recycling refills it; above that level it does not, and pollution keeps rising even if releases stop. That threshold in the lake's phosphorus level is the tipping point. Fixing scenarios makes the oracle deterministic while
retaining scenario-averaged variability in the physical response. Let
$\bar P_t=100^{-1}\sum_sP_t^{(s)}$. The first intermediate response contains
20 five-year block peaks,
\begin{equation}
 h_{1,b}(\mathbf{x})=\frac1{2.4}
 \max_{t=5(b-1),\ldots,5b-1}\bar P_t,
\quad b=1,\ldots,20.
\end{equation}
The second intermediate response is normalized discounted utility,
\begin{equation}
 h_{2,1}(\mathbf{x})=\frac{\sum_{t=0}^{99}0.98^tu_t}
 {0.1\sum_{t=0}^{99}0.98^t}.
\end{equation}
Thus $\mathbf{h}_1(\mathbf{x})=(h_{1,1},\ldots,h_{1,20})$ and
$\mathbf{h}_2(\mathbf{x})=(h_{2,1})$. The known mappings, minimized
objectives, and reference point are
\begin{equation}
 g_1(\mathbf{y})=\max_{b=1,\ldots,20}y_b,\qquad
 g_2(\mathbf{y})=1-y_1,\qquad
 f_i(\mathbf{x})=g_i(\mathbf{h}_i(\mathbf{x})),\qquad
 \mathbf{r}=(1.05,1.05).
 \label{eq:lake-objectives}
\end{equation}
This is a deterministic two-objective slice of the shallow-lake dynamics of
Carpenter et al.\ \cite{carpenter1999lake} as used in multi-objective robustness
studies \cite{ward2015tipping}; it is not the original four-objective stochastic
benchmark.  The last decision $u_{99}$ contributes to utility but, following
the implemented indexing, not to the 100 recorded lake states.  The composite
representation retains which five-year period controls water quality before
the outer maximum discards that information.
\subsection{RGB-selective multilayer nanoparticle: $d=6$, $K=6$, $N=3$}
\label{app:rgb}

Light partially reflects at every boundary between two different
materials, so a particle whose layers alternate between two materials produces
several reflections that interfere, reinforcing at some wavelengths and
canceling at others.  Choosing the layer thicknesses therefore chooses which
colors the particle scatters, and the goal is a particle that scatters one
target color band and little else. This electromagnetic design problem chooses a core radius and five concentric
shell thicknesses. For normalized $\mathbf{x}\in[0,1]^6$, the implemented dimensions are
$t_j(\mathbf{x})=30+40x_j$ nm. The layers alternate between silica, with
relative permittivity $2.04$, and a wavelength-dependent titanium-dioxide
model. For wavelength $\lambda$ measured in nanometers,
\begin{equation}
\epsilon_{\mathrm{TiO_2}}(\lambda)
=
5.913+\frac{0.2441}{10^{-6}\lambda^2-0.0803}.
\end{equation}
This formula returns the relative permittivity of titanium dioxide at
wavelength $\lambda$, which increases toward the blue end of the spectrum. The surrounding medium has relative permittivity $1.77$. The simulator evaluates
the concentric-sphere Mie calculation for orders 1--12 and both polarizations on
the discrete wavelength grid
\begin{equation}
\Lambda_s=\{350+2m:m=0,\ldots,200\}\ {\rm nm}.
\end{equation}
Mie theory gives the exact scattering from a layered sphere as an
infinite series, which the implementation truncates at twelve orders. Given
the layer thicknesses $t_j(\mathbf{x})$, the layer and medium permittivities,
and the wavelength grid $\Lambda_s$, the simulator returns a scattering value
at each wavelength. The design enters through the layer radii
$r_j(\mathbf{x})=\sum_{k\leq j}t_k(\mathbf{x})$, which determine the interference conditions at each
interface. The resulting scattering spectrum
$S\!\left(\lambda;\mathbf{r}(\mathbf{x}),\boldsymbol{\epsilon}(\lambda)\right)$
therefore depends on the design through the interface radii
$\mathbf{r}(\mathbf{x})=(r_1(\mathbf{x}),\ldots,r_6(\mathbf{x}))$ and on the
material properties through
$\boldsymbol{\epsilon}(\lambda)=(\epsilon_1(\lambda),\ldots,\epsilon_7(\lambda))$.
Specifically,
$\epsilon_1=\epsilon_3=\epsilon_5=2.04$ for silica,
$\epsilon_2(\lambda)=\epsilon_4(\lambda)=\epsilon_6(\lambda)
=\epsilon_{\mathrm{TiO_2}}(\lambda)$, and
$\epsilon_7=1.77$ represents the surrounding medium; $r_j(\mathbf{x})$
is the boundary between regions $j$ and $j+1$.
It is normalized by the particle cross-sectional area
$\pi r_6(\mathbf{x})^2$.
For brevity, we denote this spectrum by $S(\lambda;\mathbf{x})$ below.
For target bands $B_1=[400,500)$, $B_2=[500,600)$, and
$B_3=[600,700)$ nm, the implemented in-band and out-of-band responses are the
discrete sums
\begin{equation}
I_b(\mathbf{x})
=
\sum_{\lambda\in\Lambda_s\cap B_b}
S(\lambda;\mathbf{x}),
\qquad
O_b(\mathbf{x})
=
\sum_{\lambda\in\Lambda_s\setminus B_b}
S(\lambda;\mathbf{x}).
\end{equation}
Here $I_b$ is the scattering that lands inside the target band and
$O_b$ is everything else, so the objective below is the fraction of the
particle's scattering that misses the target band, and smaller values mean a
more color-selective particle. The intermediate response, known mapping, and minimized objective for band $b$ are
\begin{equation}
 \mathbf{h}_b(\mathbf{x})=\big(I_b(\mathbf{x}),O_b(\mathbf{x})\big),\qquad
 g_b(\mathbf{y})=\frac{y_2}{y_1+y_2},\qquad
 f_b(\mathbf{x})=g_b(\mathbf{h}_b(\mathbf{x})),\quad b=1,2,3.
 \label{eq:nanoparticle}
\end{equation}
The particle parameterization and scattering simulator follow Kim
et al.~\cite{kim2022deep}; the three RGB-selectivity objectives are constructed
in this work. When $g_b$ is applied to GP posterior samples, the implementation
clamps both components to be nonnegative and floors their sum at $10^{-12}$.
Each objective therefore owns a disjoint component pair.  The ratio is
nonlinear and scale-invariant, but retaining the three $\mathbf{h}_b$ doubles the independent output
models from three to six without exposing a component shared across objectives.
The reference point is $(2.5,2.5,2.5)$.
\subsection{Summit SNAr reaction: $d=4$, $K=8$, $N=2$}
\label{app:snar}
This case implements the mechanistic reaction benchmark distributed with
Summit~\cite{felton2021summit}. The \emph{starting material} is a molecule carrying two
replaceable fluorine atoms, and the reagent pyrrolidine displaces one of them.
Because either fluorine can be replaced, the reaction yields the desired
\emph{product}, a \emph{variant} in which the substitution occurred at the other
position, and---if a second pyrrolidine reacts---a \emph{byproduct} with both
fluorines replaced.

Four normalized decisions set how long the mixture stays in the reactor,
$\tau \in [0.5, 2.0]$ min; how much pyrrolidine is supplied per unit of starting
material, $e \in [1,5]$ moles per mole; the concentration of starting material
entering the reactor, $c_{\mathrm{in}} \in [0.1, 0.5]$ mol\,L$^{-1}$; and the
temperature, $T \in [30, 120]\,^\circ$C.

The state $C = (C_0, \dots, C_4)$ holds the molar concentrations of starting
material, pyrrolidine, product, variant, and byproduct, with molar masses
$M = (159.09, 71.12, 210.21, 210.21, 261.33)$ g\,mol$^{-1}$. From
$C(0) = (c_{\mathrm{in}}, e\,c_{\mathrm{in}}, 0, 0, 0)$, a fixed $256$-step RK4
solve integrates the five-species plug-flow model (a tube reactor without
back-mixing) over the residence time.
The four rate constants follow the Arrhenius law,
$k_j = 0.6\,k_j^{\mathrm{ref}} \exp[-E_j R^{-1}(T_{\mathrm{K}}^{-1} -
T_{\mathrm{ref}}^{-1})]$, with activation energy $E_j$ in kJ\,mol$^{-1}$, gas
constant $R = 8.314 \times 10^{-3}$ kJ\,mol$^{-1}$K$^{-1}$,
$T_{\mathrm{K}} = T + 273.71$, and $T_{\mathrm{ref}} = 363.71$ K; the
$(k^{\mathrm{ref}}, E)$ pairs for $a, b, c, d$ are $(57.9, 33.3)$,
$(2.70, 35.3)$, $(0.865, 38.9)$, and $(1.63, 44.8)$. With
\begin{equation}
  r_a = k_a C_0 C_1, \qquad r_b = k_b C_0 C_1, \qquad
  r_c = k_c C_1 C_2, \qquad r_d = k_d C_1 C_3,
\end{equation}
the concentration derivatives are
\begin{equation}
\begin{aligned}
  \dot C_0 &= -r_a - r_b, &\qquad \dot C_1 &= -r_a - r_b - r_c - r_d, \\
  \dot C_2 &= r_a - r_c, &\qquad \dot C_3 &= r_b - r_d, \\
  \dot C_4 &= r_c + r_d. &&
\end{aligned}
\end{equation}
Let $F(\mathbf{x})=5/\tau(\mathbf{x})$ mL\,min$^{-1}$ be the flow through
the $5$ mL reactor, and let $\mathbf{C}^{\mathrm{out}}(\mathbf{x})$ denote
the outlet concentrations produced by the RK4 integration. The retained
intermediate responses are
\begin{equation}
 \mathbf{h}_1(\mathbf{x})
 =
 \bigl(C^{\mathrm{out}}_2(\mathbf{x}),F(\mathbf{x})\bigr),
 \qquad
 \mathbf{h}_2(\mathbf{x})
 =
 \bigl(C^{\mathrm{out}}_0(\mathbf{x}),\ldots,
 C^{\mathrm{out}}_4(\mathbf{x}),F(\mathbf{x})\bigr).
\end{equation}
Thus product concentration and flow occur in both intermediate groups and are
modeled twice. For $\mathbf{y}=(y_1,y_2)\in\mathbb{R}^2$, the first known
mapping is one minus normalized \emph{space-time yield},
\begin{equation}
 g_1(\mathbf{y})
 =
 1-\operatorname{clip}\!\left(
 \frac{12M_2y_1y_2}{13000},0,1
 \right).
\end{equation}
For $\mathbf{y}=(y_1,\ldots,y_6)\in\mathbb{R}^6$, define
\begin{equation}
 W(\mathbf{y})
 =
 M_0y_1+M_1y_2+M_3y_4+M_4y_5,
\end{equation}
where $W$ is the outlet waste-mass concentration. The E-factor is
\begin{equation}
 E(\mathbf{y})
 =
 \frac{0.789y_6+10^{-3}y_6W(\mathbf{y})}
 {10^{-3}M_2y_3y_6},
 \qquad
 g_2(\mathbf{y})=\frac{\min\{E(\mathbf{y}),1000\}}{500}.
\end{equation}
The final objectives are
$f_i(\mathbf{x})=g_i(\mathbf{h}_i(\mathbf{x}))$.
The implementation clamps modeled concentrations to nonnegative values,
floors modeled flow at $10^{-8}$, and floors the E-factor denominator at
$10^{-12}$ before applying these mappings.

The composition is nonlinear and mechanistic but offers the surrogate little:
both outer maps are smooth over the attained states, product concentration and
flow are modeled twice, and flow does not affect $f_2$ at all. The reference
point is $(2.5, 2.5)$.

\subsection{RCM40 power flow: $d=34$, $K=29$, $N=2$}

RCM40 is the two-objective optimal-power-flow problem from the real-world constrained multi-objective suite \cite{kumar2021benchmark}. It involves setting the voltages on an electrical grid so that both wasted power and reactive loading are minimized. We only use the implemented hypercube constraints and not the original inequality constraints.

The grid is the standard IEEE-14 test network; its 14 nodes are called
\emph{buses}. The optimizer receives $\mathbf{x}\in[0,1]^{34}$. Its first 26
coordinates are mapped affinely to $[-1,1]$ and used as the real and imaginary
parts of the voltages at buses $2$--$14$. Bus $1$ is the \emph{slack bus} and
is fixed at $V_1=1$. The final eight coordinates are not read by the
implemented oracle, leaving 26 active dimensions.

The admittance matrix $Y = G + \mathrm{i}B \in \mathbb{C}^{14 \times 14}$ encoding how strongly each pair of buses is electrically connected, is known in advance. The real matrices $G$ and $B$ contain the
conductive and reactive parts of that connection. Multiplication yields the per-bus current and power vectors $I, S \in \mathbb{C}^{14}$ from the voltage vector $V \in \mathbb{C}^{14}$:
\begin{equation}
  I = YV,
  \qquad
  S = V \odot \overline{I},
  \qquad
  P = \Re S,
  \qquad
  Q = \Im S,
\end{equation}
where $\odot$ denotes entrywise multiplication,
$\overline{\,\cdot\,}$ denotes complex conjugation, and $\Re$ and $\Im$
return the real and imaginary parts.
Thus $P\in\mathbb{R}^{14}$ and $Q\in\mathbb{R}^{14}$ are the real- and
reactive-power injections at the buses in per-unit coordinates.

The intermediate quantities we retain and their known maps are
\begin{equation}
\label{eq:rcm40}
\begin{aligned}
  \mathbf{h}_1(\mathbf{x}) &= (P_1, \dots, P_{14}),
  &\quad
  g_1(\mathbf{y}) &= \sum_{j=1}^{14} y_j, \\
  \mathbf{h}_2(\mathbf{x}) &= (\Im I_2,\, Q_1, \dots, Q_{14}),
  &\quad
  g_2(\mathbf{y}) &= -y_1 + \sum_{j=3}^{15} y_j, \\
  f_i(\mathbf{x}) &= g_i\bigl(\mathbf{h}_i(\mathbf{x})\bigr), & i &= 1, 2
\end{aligned}
\end{equation}
Here $g_1$ sums the real-power injections, giving total real-power loss. The
second mapping gives $f_2=-\Im I_2+\sum_{j=2}^{14}Q_j$; the retained value
$Q_1$ receives zero weight. Both mappings are linear with constant Jacobians.
Their coefficients are all one for $g_1$ and are $-1$, $0$, or $1$ for $g_2$.
The reference point is $(200,500)$.
\section{Considered MOBO methods}
\label{app:methods}
For objectives $f_i(\mathbf{x})=g_i(\mathbf{h}_i(\mathbf{x}))$,
let $X_t=(\mathbf{x}_1,\ldots,\mathbf{x}_t)$ denote the evaluated designs,
$F_t=(\mathbf{f}(\mathbf{x}_j))_{j=1}^t$ their objective observations,
and $H_t=(\mathbf{h}(\mathbf{x}_j))_{j=1}^t$ the retained scalar
intermediate responses, collected in $\mathbf{h}$. The two modeling
approaches differ in the posterior samples supplied to the acquisition rule.
\emph{BlackBox-GP} fits independent GPs to the final objectives, giving samples
\begin{equation}
\widetilde f_i(\cdot)\sim
p_i\!\left(f_i(\cdot)\mid X_t,F_{t,i}\right),
\qquad i=1,\ldots,N.
\end{equation}
\emph{Composite} instead fits independent GPs to the $K$ retained scalar
intermediate responses and propagates their posterior samples through the
known mappings:
\begin{equation}
\widetilde h_k(\cdot)\sim
p_k\!\left(h_k(\cdot)\mid X_t,H_{t,k}\right),
\quad k=1,\ldots,K,
\qquad
\widetilde f_i(\mathbf{x})=
g_i\!\left(\widetilde{\mathbf{h}}_i(\mathbf{x})\right).
\end{equation}
Here $F_{t,i}$ and $H_{t,k}$ denote the corresponding observed output
columns, and $\widetilde{\mathbf{h}}_i$ collects the responses used by
$g_i$. Shared or duplicated responses are retained as specified in
Appendix~\ref{app:benchmarks}. The induced Composite objective posterior
need not be Gaussian. Both variants use the same acquisition rule and
observed objectives to measure improvement; MORBO applies these modeling
choices locally within trust regions. All definitions below use the
minimization convention.

\subsection{Log Expected Hypervolume Improvement (LogEHVI)~\cite{daulton2020differentiable,ament2023unexpected}}
\label{app:logehvi}

Expected hypervolume improvement (EHVI)
\cite{daulton2020differentiable} selects designs according to their
expected contribution to the hypervolume of the observed Pareto front.
Let $\mathcal{P}$ denote the nondominated objective vectors observed
so far and $\mathbf{r}$ the fixed reference point. For our minimization
problems, the hypervolume improvement from evaluating $\mathbf{x}$ is
\begin{equation}
    \Delta\mathrm{HV}(\mathbf{x})
    =
    \mathrm{HV}\!\left(
        \mathcal{P}\cup\{\mathbf{f}(\mathbf{x})\};\mathbf{r}
    \right)
    -
    \mathrm{HV}(\mathcal{P};\mathbf{r}).
\end{equation}
EHVI averages this improvement over the posterior distribution of
$\mathbf{f}(\mathbf{x})$. Its logarithmic counterpart targets
\begin{equation}
    \alpha_{\mathrm{LogEHVI}}(\mathbf{x})
    =
    \log \mathbb{E}\!\left[\Delta\mathrm{HV}(\mathbf{x})\right].
\end{equation}
Taking the logarithm preserves the maximizers whenever EHVI is positive.
However, directly computing EHVI and then taking its logarithm can
suffer from numerical underflow and vanishing gradients.
We therefore use BoTorch's qLogEHVI implementation
\cite{ament2023unexpected}, which evaluates a numerically stable,
smooth approximation in log space. We set $q=1$, selecting one design
at each iteration.

\subsection{Smooth Tchebycheff scalarization (STCH)~\cite{lin2024smooth,pires2026composite}}
\label{app:stch}
Smooth Tchebycheff scalarization combines the $N$ objectives into the
single scalar objective $s_\mu(\mathbf{f}(\mathbf{x});\mathbf{w})$, where
$\mathbf{w}$ specifies the tradeoff among the objectives.
For weights $w_i\geq0$ with $\sum_{i=1}^N w_i=1$, reference vector
$\mathbf{z}^\ast$, and temperature $\mu>0$, define
\begin{equation}
s_\mu(\mathbf{f}(\mathbf{x});\mathbf{w})
=\mu\log\!\left[
\sum_{i=1}^{N}\exp\!\left(
\frac{w_i(f_i(\mathbf{x})-z_i^\ast)}{\mu}
\right)\right].
\label{eq:stch-scalarization}
\end{equation}
Let $s_{\mathbf{w}}^{\mathrm{best}}$ be the smallest scalarized value
observed in the branch for weight $\mathbf{w}$. Its LogEI acquisition targets
\begin{equation}
\alpha_{\mathrm{LogEI}}(\mathbf{x};\mathbf{w})
=\log\mathbb{E}\!\left[
\left(s_{\mathbf{w}}^{\mathrm{best}}
-s_\mu(\widetilde{\mathbf{f}}(\mathbf{x});\mathbf{w})\right)_+
\right],
\label{eq:stch-logei}
\end{equation}
where $(a)_+=\max\{a,0\}$ and the expectation is over the BlackBox-GP or
Composite objective posterior. Maximizing this acquisition selects the
next design for that weight.
All STCH variants use temperature $\mu=0.05$ and five fixed weights. For two
objectives, $w_1$ is equally spaced from $0.05$ to $0.95$ and $w_2=1-w_1$; for
more objectives, deterministic normalized exponential draws cover the simplex.
The five scalarizations with different weights share one Sobol initial design,
evaluated only once, and then evolve independently. The BlackBox-GP STCH variant
applies the negative of the scalarization in Eq.~\eqref{eq:stch-scalarization},
$-s_\mu(\widetilde{\mathbf{f}};\mathbf{w})$, to each objective posterior sample
$\widetilde{\mathbf{f}}$. The Composite STCH variant first constructs
$\widetilde{\mathbf{f}}=
\bigl(g_i(\widetilde{\mathbf{h}}_i)\bigr)_{i=1}^{N}$ and then applies the same
$-s_\mu(\widetilde{\mathbf{f}};\mathbf{w})$. Both variants maximize qLogEI ($q=1$)~\cite{ament2023unexpected}
relative to the best scalarized value observed for that weight vector.
Low-dimensional branches receive eight adaptive evaluations each, while the
high-dimensional STCH branches receive twenty. For reporting only, their points
are interleaved in round-robin order so the HV trace does not depend on an
arbitrary ordering of the five weights.

\subsection{Spherical-linear STCH (Sp-STCH)~\cite{doumont2026linear}}
\label{app:spherical}

Sp-STCH retains the STCH weights, independent branches, composition, and qLogEI
steps above, but replaces every standard output GP with the following
spherical-linear model. For each output, the implementation centers $\mathbf{x}$ at $(1/2,\ldots,1/2)$, divides
coordinates by learned ARD scales, and applies a learned global radial scale
before constructing
\begin{equation}
  P(\mathbf{u})=\left[\sqrt{b_0},\;\sqrt{b_1}\frac{\mathbf{u}}{\lVert \mathbf{u}\rVert_2}\right],
  \qquad
  k(\mathbf{u},\mathbf{u}')=P(\mathbf{u})^\top P(\mathbf{u}'),
  \label{eq:spherical}
\end{equation}
with the zero-radius direction defined as zero.  A softmax constrains
$b_0+b_1=1$ and $b_0,b_1>0$.  The GP has a constant mean, a learned Gaussian
likelihood with a log-normal noise prior, and standardized outputs, following the original spherical-linear
implementation \cite{doumont2026linear}.
Posterior objective samples from these models are scalarized using
Eq.~\eqref{eq:stch-scalarization}, and the next design maximizes the
LogEI acquisition in Eq.~\eqref{eq:stch-logei}. Thus, only the surrogate
changes relative to STCH.

\subsection{Multi-Objective Regional Bayesian Optimization (MORBO)~\cite{daulton2022morbo}}
\label{app:morbo}

MORBO maintains multiple adaptive trust regions $\mathcal{T}_\ell
\subseteq\mathcal{X}$ and fits local GP models using observations near
each region. For a candidate pool $\mathcal{C}_\ell\subset\mathcal{T}_\ell$,
a Thompson sample $\widetilde{\mathbf{f}}_\ell$ assigns a sampled
objective vector to each candidate. The sampled hypervolume improvement
relative to a current nondominated set $\mathcal{P}$ is
\begin{equation}
\Delta\widetilde{\mathrm{HV}}_\ell(\mathbf{x};\mathcal{P})
=\mathrm{HV}\!\left(
\mathcal{P}\cup\{\widetilde{\mathbf{f}}_\ell(\mathbf{x})\};\mathbf{r}
\right)-\mathrm{HV}(\mathcal{P};\mathbf{r}),
\qquad \mathbf{x}\in\mathcal{C}_\ell.
\label{eq:morbo-sampled-hvi}
\end{equation}
Coordinated batch selection uses these sampled improvements across
regions, accounting for candidates already selected for the batch.
Successful regions expand, while unsuccessful regions shrink or restart.

The BlackBox-GP MORBO variant fits local GPs to final objectives, whereas the Composite MORBO variant fits local GPs
to the retained scalar entries of $\mathbf{h}_i(\mathbf{x})$ and maps each
posterior sample $\widetilde{\mathbf{h}}_i$ through $g_i$ before scoring candidate
hypervolume improvement. Both variants begin
at the same points and use the same region state, shared past observations, and batch size.
There are five trust regions, each initialized with side length $0.8$, minimum
$0.01$, and maximum $1.6$.  Candidate pools contain 512 raw points.  Each local
output GP uses a Mat\'ern-$5/2$ ARD kernel with length scales constrained to
$[0.05,4]$, output scale, and a Gaussian likelihood with a Gamma noise prior. Candidates are then selected via lower-tail winsorization and coordinated Thompson Sampling/LogEHVI.

\section{Optimization configurations}
\label{app:configuration}

\paragraph{Protocol.}
Each matched BlackBox-GP/Composite pair on the benchmarks uses the same
trial seed and an identical
scrambled-Sobol initial design, and every benchmark comparison is
repeated over $10$
paired trials, with $80$ trials for Synthetic 1. One evaluation at a design returns all objectives, or all
retained components, at once; applying the known composition to those components
does not consume a further evaluation. Hypervolume is recomputed after each
recorded design against the benchmark's fixed reference point.

\paragraph{Models.}
All models are double-precision BoTorch and GPyTorch GPs~\cite{balandat2020botorch}. Standard and spherical models are fit with \texttt{fit\_gpytorch\_mll} using exact or
summed marginal log likelihoods; the configured kernel and likelihood priors
enter that same fitting objective.

\paragraph{Surrogate models.}
The LogEHVI and STCH output GPs use a scaled Mat\'ern~5/2 kernel with ARD
in the controlled experiments and a dimension-scaled RBF kernel with ARD in
the scientific-design benchmarks. Sp-STCH and MORBO set their covariance
explicitly, as Appendices~\ref{app:spherical} and~\ref{app:morbo} describe.
Table~\ref{tab:software} gives both configurations; absolute gains should be
compared within a setting rather than across the two.

\begin{table}[h]
\centering\small
\caption{Surrogate and software configuration for the LogEHVI and STCH
output GPs in each setting. Both fit hyperparameters by
\texttt{fit\_gpytorch\_mll} in double precision with a \texttt{Standardize}
outcome transform.}
\label{tab:software}
\begin{tabular}{lll}
\toprule
 & Controlled experiments & Scientific-design benchmarks \\
\midrule
BoTorch              & 0.9.5  & 0.18.0 \\
GPyTorch             & 1.11   & 1.15.2 \\
PyTorch              & 2.11.0 & 2.12.0 \\
\addlinespace
Covariance           & \texttt{ScaleKernel(Matern 5/2)}, ARD & \texttt{RBFKernel}, ARD, no outputscale \\
Lengthscale prior    & $\mathrm{Gamma}(3.0,\,6.0)$ & $\mathrm{LogNormal}\!\left(\sqrt{2}+\tfrac{1}{2}\log d,\ \sqrt{3}\right)$ \\
Lengthscale constraint & --- & $\geq 0.025$ \\
Outputscale prior    & $\mathrm{Gamma}(2.0,\,0.15)$ & --- \\
\bottomrule
\end{tabular}
\end{table}

\paragraph{Evaluation budgets.}
Table~\ref{tab:budgets} gives the budget for each regime. The scalarization
methods spend their adaptive budget across five fixed weight vectors rather than
in one sequence, which is why the per-weight counts multiply out to the same
totals as the sequential methods.

\begin{table}[h]
\centering
\caption{Initial design size and adaptive evaluation budget per trial. The
high-dimensional regime covers L--A, Lake, and RCM40.}
\label{tab:budgets}
\begin{tabular}{llccl}
\toprule
Regime & Method & Initial & Adaptive & Adaptive budget spent as \\
\midrule
Low-dimensional  & LogEHVI & 5  & 40  & sequential, one candidate per step \\
                 & STCH    & 5  & 40  & 5 weights $\times$ 8 evaluations \\
\addlinespace
High-dimensional & Sp-STCH & 20 & 100 & 5 weights $\times$ 20 evaluations \\
                 & MORBO   & 20 & 100 & 20 batches $\times$ 5 candidates \\
\bottomrule
\end{tabular}
\end{table}

\paragraph{Acquisition optimization.}
LogEHVI, STCH, and Sp-STCH maximize their acquisition function with BoTorch's
\texttt{optimize\_acqf}, which seeds a multistart search from raw Sobol
candidates and refines it with SciPy's L-BFGS-B~\cite{balandat2020botorch}.
LogEHVI and STCH use 128 raw candidates with 8 restarts, and Sp-STCH uses 256 raw candidates
with 10 restarts. MORBO does not use L-BFGS-B: it sequentially selects twenty batches of
five candidates from a discrete Sobol pool of 512 points drawn inside its trust
regions.



\end{document}